\documentclass{article}

\usepackage[preprint, nonatbib]{neurips_2022}

\usepackage[utf8]{inputenc} % allow utf-8 input
\usepackage[T1]{fontenc}    % use 8-bit T1 fonts
\usepackage{bbm}
\usepackage{hyperref}       % hyperlinks
\usepackage{url}            % simple URL typesetting
\usepackage{booktabs}       % professional-quality tables
\usepackage{amsfonts}       % blackboard math symbols
\usepackage{nicefrac}       % compact symbols for 1/2, etc.
\usepackage{microtype}      % microtypography
\usepackage{amsmath}
\usepackage{amsthm}
\usepackage{mathtools}
\usepackage[backend=bibtex]{biblatex}
\usepackage{graphicx}
\usepackage{subcaption}
\usepackage{tikz}
\usetikzlibrary{bayesnet}
\usetikzlibrary{arrows, arrows.meta, shapes, calc, intersections, positioning, decorations.pathreplacing, fit, backgrounds, patterns}

\theoremstyle{definition}

\newcommand{\cU}{\mathcal{U}}

\newcommand{\cP}{\mathcal{P}}

\newcommand{\cL}{\mathcal{L}}

\newcommand{\bx}{\boldsymbol x}
\newcommand{\by}{\boldsymbol y}

\newcommand{\bxi}{\boldsymbol \xi}

\newcommand{\bE}{\mathbb{E}}
\newcommand{\bR}{\mathbb{R}}

\newcommand{\bsigma}{\bar{\sigma}}

\newcommand{\diag}{\text{diag}}

\title{Uncertainty Calibration in Bayesian Neural Networks via Distance-Aware Priors}

\author{%
	Gianluca Detommaso\thanks{equal contribution}\\
	AWS\\
	\texttt{detomma@amazon.de}
\And
	Alberto Gasparin\footnotemark[1]\\
	Amazon\\
	\texttt{albgas@amazon.de}
\And
	Andrew Wilson\\
	Amazon\\
	\texttt{wilsmman@amazon.com}
\And
	Cedric Archambeau\\
	AWS\\
	\texttt{cedrica@amazon.de}
}

\bibliography{biblio}

\begin{document}
\maketitle

\begin{abstract}
As we move away from the data, the predictive uncertainty should increase, since a great variety of explanations are consistent with the little available information. We introduce Distance-Aware Prior (DAP) calibration, a method to correct overconfidence of Bayesian deep learning models outside of the training domain. We define DAPs as prior distributions over the model parameters that depend on the inputs through a measure of their distance from the training set. DAP calibration is agnostic to the posterior inference method, and it can be performed as a post-processing step. We demonstrate its effectiveness against several baselines in a variety of classification and regression problems, including benchmarks designed to test the quality of predictive distributions away from the data.
\end{abstract}

\section{Introduction}\label{sec:intro}
Uncertainty quantification is crucial in a variety of real-world applications, such as human decision making in medicine, risk assessment in automatic vehicles, anomaly detection in fraud and cybersecurity. Uncertainty is often compartmentalized into aleatoric and epistemic components \cite{kendall2017uncertainties, malinin2018predictive, hullermeier2021aleatoric}. Aleatoric uncertainty characterizes the intrinsic stochasticity in the data given the data generating process. Epistemic uncertainty describes our inability to identify the data generating process because of lack of information \cite{hullermeier2021aleatoric}. Aleatoric uncertainty is irreducible because, given a data generating process, the noise in the data cannot be decreased. Epistemic uncertainty can be reduced by either better modelling or by collecting more data. In general, we expect epistemic uncertainty to be small close to the training set, where data is available, and large outside of the training domain, when data is either not available or very sparse. We refer to these two situations as in-distribution and out-of-distribution (OOD). Deep learning models tend to overfit the training data, erroneously providing the same uncertainty regardless of where we wish to make a prediction \cite{guo2017calibration}.

In Bayesian neural networks, one places a prior over parameters, inducing a prior over functions that represents epistemic uncertainty. The most commonly used prior is an independent Gaussian prior over parameters \cite{wilson2020bayesian}. In practice issues with approximate inference can have a significant effect on the ability of these approaches to represent growing uncertainty away from the data \cite{maddox2019simple, foong2019between, yao2019quality}. Moreover, even with high quality inference, these priors lead to poor predictive distribution in OOD settings \cite{izmailov2021dangers}.

% Instead, 
In the classic framework,  model parameters and input data are assumed to be independent \textit{a priori} \cite{wilson2020bayesian}, failing to represent our honest beliefs that the uncertainty should grow as we move away from the data. However, representing these beliefs is particularly desirable in OOD settings, where the data provides none or scarce information about the model parameters, and the prior distribution is the main determinant of the epistemic uncertainty. 

%Because approximations of the posterior distribution harm the possibility to be more uncertain OOD even after the training targets are observed \cite{yao2019quality}, input-dependent prior distributions may be useful to mitigate overconfidence OOD.

%leading to erroneously conclude that the learned mapping between input and target data holds similarly elsewhere. It follows the epistemic uncertainty is often poorly represented, especially in OOD settings \cite{guo2017calibration, foong2019between, izmailov2021dangers}.

%In the classic Bayesian deep learning, %framework,  model parameters and input data are assumed to be independent \textit{a priori} \cite{wilson2020bayesian} (this is not the case in amortized VI \cite{zhang2018advances}). Thus, before the target data is observed, the model uncertainty does not vary over different inputs. In particular, a priori it is not possible to convey %the reasonable intuition that the model should be more uncertain when evaluated at inputs less similar to the ones in the training data. 

For this reason, we drop the independence assumption between model parameters and inputs, and propose uncertainty calibration via \textit{distance-aware priors} (DAPs). These priors over the model parameters induce larger model uncertainty at inputs further from the training data. Since the prior distribution has a strong influence on the predictive distribution in parts of the input space where data is sparse or not available, we can calibrate the input-dependent increase in prior uncertainty to remedy overconfidence OOD. As we do not impose any restriction on the target data or likelihood function, DAP calibration can be performed in both classification and regression problems. Furthermore, DAP calibration is agnostic to the posterior inference method \cite{lakshminarayanan2017simple, gal2016dropout, wilson2016deep, maddox2019simple, osawa2019practical, zhang2019cyclical}, and it can be performed as a post-processing step. 

We remark that although DAP calibration only targets OOD uncertainty, it can be seamlessly combined with methods better tailored to calibrate in-domain uncertainty, such as temperature scaling \cite{platt1999probabilistic, guo2017calibration, kuleshov2018accurate, laves2020well, vovk2020conformal, kapoor2022uncertainty}. Nevertheless, throughout this paper we will exclusively focus on OOD uncertainty calibration, in order to avoid confounding results.

% We introduce a simple distance-aware Gaussian prior, capturing the dependence of the model parameters to the inputs through a parametrized distance. We propose a distance measure based on neural network feature mapping, and propose a calibration loss function that allows to achieve a desired level of uncertainty OOD, in both classification and regression problems. We study the effectiveness of DAP calibration over several experiments, and against several methods for OOD uncertainty estimation.

We can summarize the contributions of this work as follows. (1) We propose a novel method to calibrate OOD uncertainty estimates of Bayesian neural network models. (2) We introduce a simple distance-aware Gaussian prior, capturing the dependence of the model parameters to the inputs through a parametrized distance. (3) We consider a distance measure based on neural network feature mapping and (4) introduce a calibration loss function to achieve a desired level of uncertainty OOD, in both classification and regression problems. (5) We propose an importance sampling procedure that allows to perform calibration as a post-processing step. (6) We study the effectiveness of DAP calibration over several experiments, and against several methods for OOD uncertainty estimation.

% We can summarize the contributions of this work as follows: (1) We propose a novel method for improving uncertainty calibration in Bayesian neural networks making the predictive distribution input distance-aware (2)  (3) To allow our approach to scale we propose to use importance sampling to estimate the posterior expectation of any statistics, which is particularly relevant for estimating the posterior predictive distribution (4) The method we propose does not require any change in the training procedure and can thus be applied to any pretrained Bayesian model (5) We show that our method is competitive with popular methods for uncertainty estimation as well as methods specialised in out-of-distribution detection.

\section{Related Work}
The concept of distant-aware uncertainty is native in Gaussian Processes (GPs) \cite{williams2006gaussian}, where a kernel captures a measure of distance between pairs of inputs. Modern approaches combine Radial Basis Function (RBF) kernels with deep feature extractors, i.e.~deep neural networks that transform the input space in order to obtain a better fit of the data \cite{chen2016deep}. This approach is commonly referred to as Deep Kernel Learning (DKL) \cite{pmlr-v51-wilson16,wilson2016stochastic}, and has recently inspired a variety of methods for deterministic uncertainty estimation. SNGP \cite{liu2020simple} is one such method, which builds upon DKL proposing to approximate a GP via Random Fourier Features (RFFs) \cite{rahimi2007random}. In the same work the authors propose a theoretical analysis of the uncertainty estimation problem, identifying input distance awareness as a key property for the task. In practice, they show how requiring deep learning models to satisfy a bi-Lipschitz condition (enforced via Spectral Normalization (SN) \cite{miyato2018spectral}) can actually make the model's predictive uncertainty input distance aware. \cite{van2021on} adopts a similar approach to the one described in SNGP, where SN is used along with residual networks to enforce the bi-Lipschitz condition, while a variational inducing point GP approximation is used instead of RFFs. 

In Deep Uncertainty Quantification (DUQ) \cite{van2020uncertainty}, another single-forward pass method, the bi-Lipschitz constraint is enforced via a two-sided gradient penalty \cite{gulrajani2017improved}, while predictions are made via a RBF network which measures the distance between the transformed test input and some class centroids. This limits its scope to classification only. These methods are competitive with Deep Ensembles on multiple OOD benchmarks, but still requires changes in the training procedure and, as such, cannot be applied as a post-processing step. A simpler alternative that addresses this issue is Deep Deterministic Uncertainty (DDU) \cite{mukhoti2021deterministic}, where Gaussian Discriminant Analysis is fit after model training and later used for characterizing epistemic uncertainty, while the softmax predictive entropy is used for aleatoric uncertainty. Similarly to \cite{van2020uncertainty,liu2020simple,van2021on}, this method performs best when SN is used in the feature extractor, outperforming both SNGP and DUQ on popular OOD classification benchmarks. 

In this work, rather than introducing a distance-aware output function, we define distance-aware prior distributions in order to obtain uncertainty estimates that grow as we move away from the training data. Like DDU, DAP calibration does not require any change to neither the deep learning model, nor the posterior inference procedure.

\section{Distance-aware priors}\label{sec:daps}

We assume to be given a training data set $\{(x_i, y_i)\}_{i=1}^N$ of inputs $x_i$ and targets $y_i$
% , and a validation data set $\{(x_i^*, y_j^*)\}_{j=1}^{N^*}$ of inputs $x_j^*$ and targets $y_j^*$. 
We denote collections of training inputs and targets by $\bx$ and $\by$, respectively. After the training data is observed, we are interested in quantifying the uncertainty of validation targets $y_j^*$, given validation inputs $x_j^*$, for $j=1,\dots,N^*$. We do so by looking at statistics defined over the predictive density $p(y_j^*|x_j^*, \bx, \by)$,  %. By marginalization, the latter can be expressed as
which is obtained by integrating out the parameters $\theta\in\bR^{n_\theta}$ of the deep learning model: 
\begin{equation}\label{eq:pred_distr}
    p(y_j^*|x_j^*,\bx,\by) = \int_{\bR^{n_\theta}} p(y_j^*|\theta,x_j^*)\,p(\theta|x_j^*,\bx,\by)\,d\theta,
\end{equation}
where %$\theta\in\bR^{n_\theta}$ denotes model parameters,
$p(y_j^*|\theta,x_j^*)$ denotes the test likelihood function and $p(\theta|x_j^*,\bx,\by)$ the posterior distribution. 

In \eqref{eq:pred_distr}, we followed the typical assumption %in the literature \cite{bishop2006pattern, hastie2009elements} 
that the test target is independent of the training data given the test input and the model parameters \cite{bishop2006pattern, hastie2009elements}, and hence we can write $p(y_j^*|\theta,x_j^*,\bx,\by)$ as $p(y_j^*|\theta,x_j^*)$. The idea behind this assumption is that the parameters $\theta$ capture all the information in the training set necessary to characterize $y_j^*$.

A second common assumption is that the prior distribution of the model parameters $\theta$ is independent of the inputs $\bx$ and $x_j^*$. In scenarios where the interpretation of the model parameters is explainable, this assumption may be justified by the lack of a clear causal relation from inputs to parameters. In %black-box 
deep learning models, on the other hand, the model parameters have no clear interpretation, hence the independence assumption is dictated by practical reasons rather than a principled understanding of the dependency structure. In fact, in order to compensate for the consequences of this independence assumption, input data are often normalized and outliers removed, with the goal of making parameters ``compatible'' with all the inputs passed to the model, instead of letting the distribution of the model parameters depend on the actual inputs. While practical, from an epistemic (model) uncertainty point of view this assumption may be harmful, as it is taking away from the model parameters the possibility to express more or less uncertainty depending on the inputs. For example, it is reasonable to assume that the model should be able to express \textit{a priori} more uncertainty far away from the training inputs, and less close by. 

\begin{figure}[t]
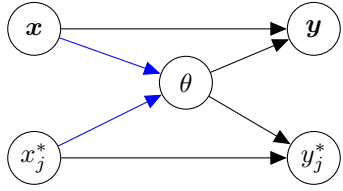

\centering
	\tikz{
		% nodes
		\node[latent] (x) {$\bx$};%
		\node[latent, below=of x] (xj) {$x_j^*$};%
		\node[latent, right=of xj, xshift=0.3cm, yshift=1cm] (t) {$\theta$};
		\node[latent, right=of x, xshift=2cm] (y) {$\by$};%
		\node[latent, right=of xj, xshift=2cm] (yj) {$y_j^*$};%
		% edges
		\edge[color=blue]{x, xj} {t};
		\edge{t} {y, yj};
		\edge{x} {y};
		\edge{xj} {yj};
	}	
	\caption{Unlike standard priors in deep learning, we assume not only the targets $y$ but also the model parameters $\theta$ explicitly depend on the data inputs $x$, as indicated by the blue arrows.}
	\label{fig:dependencies}
\end{figure}

For these reasons, %in this work 
we do not assume that the model parameters are independent of the inputs, and %, following the intuition above, we 
will construct prior distributions $p(\theta|\bxi)$ that express more uncertainty when arbitrary collections of inputs $\bxi$ are far from the training inputs, and less otherwise. We will refer to this type of distributions as \textit{distance-aware priors}. Figure \ref{fig:dependencies} shows the %graph of assumed dependencies
directed graphical model, where the blue arrows correspond to the dependencies that we account for, in contrast with the standard assumption described %above. 
earlier. 
It also makes clear that in this framework there is %generally 
a difference between the posterior density $p(\theta|x_j^*,\bx,\by)$ that appears in \eqref{eq:pred_distr} and $p(\theta|\bx,\by)$, since $\theta$ and $x_j^*$ are not independent given the training data $\bx$ and $\by$.

\subsection{Independent and joint predictions are not the same}\label{sec:indep_pred}
We note that because of the blue arrows in Figure \ref{fig:dependencies}, %are modelled, 
in this framework there is %generally 
a difference between predicting independently and jointly: the joint predictive $p(\by^*|\bx^*,\bx,\by)$ and the product of the independent predictive distributions $p(y_j^*|\bx_j^*,\bx,\by)$ are not the same, since $p(\theta|x_j^*, \bx)$ and $p(\theta|\bx^*, \bx)$ are different. 

The right distribution to use depends on the application. If we are interested in a certain prediction given that we are observing a certain set of inputs (e.g.~predicting the score of a student, given that scores are normalized over the results of the whole class), the joint predictive may be the right distribution to use. Otherwise, if we are interested in simultaneous independent predictions, we would rather use the independent predictive distribution. Throughout this paper we use the individual predictive, since we would argue that the latter situation is more commonly desired in machine learning applications.

\subsection{Distance-aware priors}\label{sec:distaware_prior}
We define distance-aware priors as distributions over the model parameters that depend on the inputs through a measure of their distance from the training inputs. In this work we use the term \textit{distance} to indicate set-to-set distance, typically from an arbitrary set of inputs $\bxi$ to the training set $\bx$. We require a distance function to be always non-negative, and to satisfy the triangular inequality. Notice that we drop the symmetry assumption since it is not needed for the purpose of this work. %, and extend the definition to include also non-symmetric distances. 
We use the notation $d_\phi(\bxi)$ to indicate a distance from $\bxi$ to $\bx$ parametrized by $\phi\in\bR^{n_\phi}$, where for sake of notation we leave the argument $\bx$ implicit.

While there are many possible ways to define a distance-aware prior, throughout this paper we will consider the following distance-aware Gaussian prior:
\begin{equation}\label{eq:da_prior}
p(\theta|\bxi) = \text{Normal}(\theta\,|0,\, (\sigma_0 + d_\phi(\bxi))^2 I_{n_\theta}),
\end{equation}
where $\bxi$ is an arbitrary collection of inputs, $\sigma_0\in\bR_+$ is a standard deviation parameter, $I_{n_\theta}$ is an identity matrix of size $n_\theta$ and $d_\phi(\bx^*)\in\bR_+$ is a distance metric parametrized by $\phi\in\bR^{n_\phi}$. 

The definition of distance-aware Gaussian in \eqref{eq:da_prior} could be easily extended by considering an array of distance $d_\phi(\bxi)\in\bR_+^{n_\theta}$ instead of a scalar quantity, which would allow to rescale the prior variance independently for each component of $\theta$. However, throughout this paper we will not consider this option, since a scalar distance is practically easier to calibrate over a low-dimensional parameter $\phi$.

\subsection{Requiring the distance to be zero at training inputs}\label{sec:dx=0}
We further require that $d_\phi(\bx)=0$, i.e.~the distance evaluated at the collection of training inputs must be 0. It follows that, at training time, the distance-aware Gaussian prior introduced in \ref{sec:distaware_prior} boils down to a classic Gaussian prior, i.e.~$p(\theta|\bx) = \text{Normal}(\theta|0, \sigma_0^2)$,
which does not depend on the distance nor on its parameters $\phi$. Furthermore, by Bayes' theorem we can write
\[ p(\theta|\bx,\by) = p(\by|\theta,\bx)p(\theta|\bx) / p(\by|\bx). \]
The posterior $p(\theta|\bx,\by)$ is not distance dependent since none of the terms on the right hand side depend on distance. As such, the latter corresponds to the posterior of standard frameworks, and it can be inferred via classic posterior inference methods \cite{neal2011mcmc, blei2017variational, mackay1992bayesian, lakshminarayanan2017simple, gal2016dropout}. The difference between classic priors and distance-aware priors arises at prediction time, when we need to evaluate the posterior $p(\theta|x_j^*,\bx,\by)$. In fact, by Bayes' theorem this posterior can be expressed in terms of the prior $p(\theta|x_j^*,\bx)$, which in turn depends on the distance $d_\phi([x_j^*, \bx])$, that is generally non-zero.

\subsection{Choice of distance in distance-aware priors}\label{sec:distance}
In this paper we %employ 
consider a distance 
\begin{equation}\label{eq:distance_decomp}
    d_\phi(\bxi) := g(\phi)d_0(\bxi),
\end{equation}
where $g(\cdot)$ is a scalar positive function of $\phi\in\bR$, while $d_0(\bxi)$ is a distance independent of $\phi$. We address $d_0(\cdot)$ as \textit{pre-distance}, since it corresponds to the distance $d_\phi(\cdot)$ before the scaling $g(\phi)$ has been applied. The choice in \eqref{eq:distance_decomp} is practically compelling for calibration over $\phi$, since after the pre-distance $d_0(\bxi)$ is pre-computed, the distance $d_\phi(\bxi)$ can be cheaply evaluated %over 
for different values of $\phi$. The function $g(\phi)$ acts as a scaling factor, which regulates the strength of the distance in the variance of the distance-aware prior, for the specific collection of inputs $\bxi$. 

%As a 
We require the scaling function $g(\phi)$ %, we want 
to be a smooth function with positive image defined %over the real domain
on the reals. This condition allows us to optimize %over 
with respect to $\phi$ in an unbounded domain. %, while %keeping 
%the distance remains positive. 
In this paper we simply choose $g(\phi)=e^\phi$, but other functions, e.g. a softplus or a scaled sigmoid, are also valid choices. We take the parameter $\phi$ to be a scalar to promote scalability and robustness; %, but 
other choices are possible.

%In this paper we study 
% We consider the two following pre-distances $d_0(\bxi)$. 

As pre-distance, we consider a projected nearest neighbor asymmetric distance given by
\begin{equation}\label{eq:d0_nn}
d_0(\bxi) := \max_k\min_{i=1\dots,N} \|\cP(\xi_k) - \cP(x_i)\|_2,
\end{equation}
where $\xi_k$ refers to the inputs in the collection $\bxi$ %, while 
and $\cP(\cdot)$ is a projection operator that transforms the input space into a low-dimensional latent space. 
% Although for low-dimensional inputs the projection may just be defined as the identity operator, in high-dimension it may %be 
% help reduce %classic 
% feature collapse \cite{aggarwal2001surprising}. 
The projector $\cP$ can be defined in several ways, for example via a variational autoencoder (VAE) \cite{kingma2013auto}, random kitchen sinks \cite{rahimi2007random}, or as a deep feature extractor, i.e.~as the forward pass of a trained deep learning model, up to the last hidden layer \cite{chen2016deep}. Because at posterior inference time we already train a deep learning model from input to output, throughout this paper we exploit the latter solution as a projector. We remark that the pre-distance in \eqref{eq:d0_nn} is strongly related to the Hausdorff distance \cite{rockafellar2009variational}, which, unlike \eqref{eq:d0_nn}, is symmetric. Since the symmetry property is not needed for the purpose of this paper, we use \eqref{eq:d0_nn} because cheaper to compute.

It is important to notice that, since we are interested in independent predictions, as discussed in Section \ref{sec:indep_pred}, we only ever need to evaluate the distance at collections $\bxi=\bx$ and $\bxi=[x_j^*,\bx]$. From \eqref{eq:d0_nn}, it is immediate to see that $d_\phi(\bx)=0$, satisfying the property required in Section \ref{sec:distaware_prior}, which makes the posterior $p(\theta|\bx,\by)$ independent of the distance. On the other hand, we observe that 
\[ d_\phi([x_j^*,\bx])=d_\phi(x_j^*) = \min_{i=1\dots,N} \|\cP(x_j^*) - \cP(x_i)\|_2. \]
Hence the maximum operator in \eqref{eq:d0_nn} is never needed, and it could be formally replaced by any other function over the index $k$, e.g.~a sum operator.   

% A second pre-distance that we consider is the projected empirical biased maximum mean discrepancy (MMD) given by
% \begin{equation}\label{eq:d0_mmd}
% d_0(\bxi) := \sqrt{\frac{1}{N_\xi^2}\sum_{i=1}^{N_\xi}\sum_{j=1}^{N_\xi}k_{\cP}(\xi_i, \xi_j) - 2\frac{1}{N_\xi N}\sum_{i=1}^{N_\xi}\sum_{j=1}^{N}k_{\cP}(x_i, \xi_j) + \frac{1}{N^2}\sum_{i=1}^N\sum_{j=1}^{N}k_{\cP}(x_i, x_j)},
% \end{equation}
% where $k_\cP(\cdot,\cdot)$ is a kernel defined as $k_\cP(\xi_i,\xi_j):=k(\cP(\xi_i),\cP(\xi_j))$, with $k(\cdot,\cdot)$ being another kernel and $\cP$ a projection operator. We use the biased empirical MMD since the term in the square-root is guaranteed to be non-negative, unlike its unbiased version \cite{gretton2012kernel}. For the purpose of this paper, employing a biased MMD estimator does not constitute a problem, while a negative distance would be. Analogously as above, we observe that we need to evaluate the distance only at $\bxi=\bx$ and $\bxi=[x_j^*,\bx]$. It is immediate to see from \eqref{eq:d0_mmd} that $d_\phi(\bx)=0$, as desired. On the other hand, when $\bxi=[x_j^*,\bx]$, the definition \eqref{eq:d0_mmd} can be simplified to
% \[ d_0([x_j^*,\bx]) = \frac{1}{N+1}\sqrt{\frac{1}{N^2}\sum_{i=1}^N\sum_{\ell=1}^N k(x_i,x_\ell) - \frac{2}{N}\sum_{i=1}^N k(x_i,x_j^*) + 1}. \] 

\section{Estimation and calibration of predictive uncertainty}
In this section we introduce importance sampling as a convenient method to estimate statistics with respect to the predictive $p(y_j^*|x_j^*, \bx, \by)$. Furthermore, we introduce a calibration strategy to correct overconfidence OOD as a post-processing step.

\subsection{Importance sampling to compute statistics over the posterior predictive distribution}
As discussed in Section \ref{sec:indep_pred}, we are interested in generating independent predictions for each of the inputs $x_j^*$. In order to do so, one may write the predictive distribution in \eqref{eq:pred_distr} as an expectation with respect to the posterior $p(\theta|x_j^*,\bx,\by)$, and estimate it via a Monte Carlo integration. However, since for each $j$ we require an individual posterior inference approximation, inference can become quickly infeasible. Furthermore, as we will see in Section \ref{sec:calib}, in order to calibrate the distance $d_\phi([x_j^*,\bx])$ in the distance-aware priors, we might want predictive estimates for several values of $\phi$, making this approach even more computationally intensive. 

In order to circumvent all these problems at once, %in this paper
we propose %a simple solution via 
to use importance sampling, where we adopt the posterior $p(\theta|\bx,\by)$ as importance distribution. We %notice 
note that the latter posterior does not depend on $x_j^*$, nor %Furthermore, it does not
does it depend on $\phi$: %, since we have assumed in Section \ref{sec:dx=0} that 
as in Section \ref{sec:dx=0}, we required $d_\phi(\bx)=0$. We then rewrite the predictive in \eqref{eq:pred_distr} as
\begin{equation}\label{eq:pred_is}
    p(y_j^*|x_j^*,\bx,\by) = \bE_{\theta|\bx,\by}[p(y_j^*|\theta,x_j^*)w(\theta|x_j^*,\bx,\by)],
\end{equation}
where we refer to $w(\theta|x_j^*,\bx,\by)$ as the \textit{posterior weight} defined by
\begin{equation}\label{eq:post_weight}
    w(\theta|x_j^*,\bx,\by):=\frac{p(\theta|x_j^*,\bx,\by)}{p(\theta|\bx,\by)} = \frac{p(\theta|x_j^*,\bx)}{p(\theta|\bx)}\frac{p(\by|\bx)}{p(\by|x_j^*,\bx)}.
\end{equation}
We provide details in Appendix \ref{app:post_weight}. Hence, we can infer the posterior $p(\theta|\bx,\by)$ a single time, generate samples from it, and estimate the predictive in \eqref{eq:pred_is} via Monte Carlo for each $x_j^*$ and each $\phi$. 

More generally, we can use %an analogous approach for 
this approach to estimate the posterior expectation of any statistics $f(\theta)$: %integrated with respect to the posterior $p(\theta|x_j^*,\bx,\by)$. We have
\begin{equation}\label{eq:pred_stats}
\bE_{\theta|x_j^*,\bx,\by}[f(\theta)] = \bE_{\theta|\bx,\by}[f(\theta)w(\theta|x_j^*,\bx,\by)].
\end{equation}
%For example, if 
If $f(\theta) := p(y_j^*|\theta,x_j^*)$, %this corresponds to 
we get
the predictive distribution in \eqref{eq:pred_is}, while if $f(\theta):=\bE_{Y_j^*|\theta,x_j^*}[Y_j^*]$, we %have the
get the predictive mean. Other statistics such as predictive variance and entropy, as well as their decomposition into aleatoric and epistemic components, can be derived analogously.

We remark once more that while it might appear unfamiliar that the posterior $p(\theta|x_j^*,\bx,\by)$ depends on the validation input $x_j^*$, this dependence is due to the model dependencies highlighted in blue in Figure \ref{fig:dependencies}, for which $x_j^*$ and $\theta$ are not independent even when $\bx$ and $\by$ are observed. Similarly, the marginal $p(\by|x_j^*,\bx)$ also depends on $x_j^*$, since $\by$ and $x_j^*$ are not independent when $\theta$ is not observed. This phenomenon has a natural analogy with the \textit{observer effect} in physics \cite{dirac1981principles}, stating that the outcome of a system can depend on the point of view of who is observing it, and be affected by the act of observing itself. In our framework, the system is the probabilistic model, and the point of view corresponds to the index $j$, for which an input $x_j$ is observed.

\subsubsection{Marginal ratio estimation and the Gaussian posterior scenario}
The computation of the posterior weight in \eqref{eq:post_weight} involves the estimation of %what we adressed as 
the marginal ratio, i.e.~the ratio between $p(\by|\bx)$ and $p(\by|x_j^*,\bx)$. %In general, each 
Each of these marginals can be estimated via %a 
Monte Carlo %approach over the marginalized formulations
integration: 
$p(\by|\bx)=\bE_{\theta|\bx}[p(\by|\theta,\bx)]$ and $p(\by|x_j^*, \bx)=\bE_{\theta|x_j^*, \bx}[p(\by|\theta,\bx)]$. 
One alternative, for which a derivation is provided in Appendix \ref{app:gauss_marginal}, is to observe that
\[ \frac{p(\by|\bx)}{p(\by|x_j^*,\bx)} = \bE_{\theta|\bx,\by}\left[\frac{p(\theta|x_j^*,\bx)}{p(\theta|\bx)}\right]^{-1}, \]
which can be again approximated via Monte Carlo with samples from the posterior $p(\theta|\bx,\by)$. If we furthermore approximate the posterior as a Gaussian distribution, which is often the case in several popular Bayesian inference methods, e.g.~\cite{blei2017variational, mackay1992bayesian}, then the marginal can be estimated in closed-form. Indeed, given the Gaussian distance-aware prior definition in \eqref{eq:da_prior}, suppose that the posterior distribution $p(\theta|\bx,\by)$ is approximated as a diagonal Gaussian with mean $\mu=[\mu_1,\dots,\mu_{n_\theta}]$ and variance $\sigma^2=[\sigma_1^2,\dots,\sigma^2_{n_\theta}]$. In addition, let us assume that $\max_i{\sigma_i^2} < \sigma_0^2$, meaning that the data information contracts the prior variance over all directions of the parameter space. Then we have
\[ \frac{p(\by|\bx)}{p(\by|x_j^*,\bx)} = \left(\frac{\sigma_0+d_j}{\sigma_0}\right)^{n_\theta} \prod_{i=1}^{n_\theta}\frac{\sigma_i}{\beta_{i,j}}\exp\left(\frac{1}{2}\frac{\mu_i^2}{\sigma_i^2}\left(1 - \frac{\beta_{i,j}^2}{\sigma_i^2}\right)\right), \]
with $d_j:=d([x_j^*,\bx])$, $\beta_{i,j}^{-2}:=\tfrac{1}{\sigma_i^2} - \tfrac{1}{\gamma_j^2}$ and $\gamma_j^{-2} := \tfrac{1}{\sigma_0^2} - \tfrac{1}{(\sigma_0+d_j^2)}$. See Appendix \eqref{app:gauss_marginal}.
	
\subsection{Calibrating the distance as a post-processing step of training}\label{sec:calib}
We have introduced distance-aware priors in order to correct overconfidence of predictions OOD. We do so by increasing the prior variance according to the distance of the validation inputs from the training set. However, for the total predictive uncertainty to be reasonable, the distance needs to be calibrated. Since we use a parametric distance $d_\phi(\cdot)$, specifically as in \eqref{eq:distance_decomp}, we calibrate the distance by picking %the 
$\phi$ such that %obtains 
we reach the minimum of a certain calibration loss. 

There are many plausible choices for the calibration loss. For example, given a validation data set, one could maximize a sum over $j$ of the test log-marginal likelihood, i.e.~$\sum_j\log p(y_j^*|x_j^*)$, in order to find the parameter $\phi$ for which the model best fits the validation data. A similar choice may involve maximizing the summed log-predictive $\sum_j\log p(y_j^*|x_j^*, \bx, \by)$, where the marginalization is with respect to the posterior rather than the prior. Alternatively, one could choose an empirical Bayes approach \cite{maritz2018empirical}. In this paper we choose the following loss function, which appeared to work best across a variety of experiments:
\begin{equation}\label{eq:calib_loss}
    \cL(\phi) = \frac{1}{N^*}\sum_{j=1}^{N^*}\|\cU_\phi(x_j^*) - \gamma\|^2.
\end{equation}
The loss in \eqref{eq:calib_loss} forces a certain measure of uncertainty $\cU_\phi(x_j^*)$ to be as close as possible to a certain %amount 
$\gamma$ at a set of validation inputs $x_j^*$, for $j=1,\dots,N^*$. For example, in a classification case, we may choose $\cU_\phi(x_j^*)$ to be the predictive mean evaluated at some OOD inputs $x_j^*$, and $\gamma=\left[\tfrac{1}{m},\dots,\tfrac{1}{m}\right]\in\bR^m$, where $m$ is the number of classes. This loss steers the model towards a state of total uncertainty at the given inputs, counteracting overconfidence OOD. In  regression, we may choose $\cU_\phi(x_j^*)$ to be the epistemic variance evaluated at some OOD inputs $x_j^*$, and $\gamma$ to be some value based on a quantile of the epistemic variances estimated at the training inputs. This strategy drives the epistemic variance to be large OOD compared to the variance in-domain, which again counteracts the overconfidence issue. 

The OOD inputs $x_j^*$ might be available for a certain test case, in which case they might as well be directly used. If these are not available, a possible strategy is to generate them from an arbitrary distribution, with a domain just outside of the training domain. With low-dimensional inputs, this is convenient, as one may be able to visualize where the training domain is located, and simulate OOD inputs accordingly. When this is less clear, a possible drawback is that if the generated inputs are very far from the training data, even after calibrating with \eqref{eq:calib_loss} the predictive uncertainty might still be too small in the nearest of the training domain. In order to avoid this issue, in high-dimensions we prefer to take a set of validation inputs on which we either misclassify, or we misregress above a certain error threshold. Intuitively, inputs on which we predict badly are often the furthest from the training domain, which we will confirm experimentally in Section \ref{sec:experiments}. 
	
\section{Experiments}\label{sec:experiments}
In this section, we first demonstrate the effect of DAP calibration on low-dimensional toy examples, in order to visualize its effect on OOD uncertainty estimates. Then we test DAP calibration on UCI Gap \cite{foong2019between}, a set of benchmarks designed to test the quality of predictive distributions away from the data. Finally, we study the ability of our method to discriminate in-distribution and OOD data, on MNIST, Fashion-MNIST and CIFAR10. We compare its performance against several baselines. 

\subsection{Toy examples}\label{sec:toy_examples}
\begin{figure}[t]
    \centering
    \includegraphics[width=\textwidth]{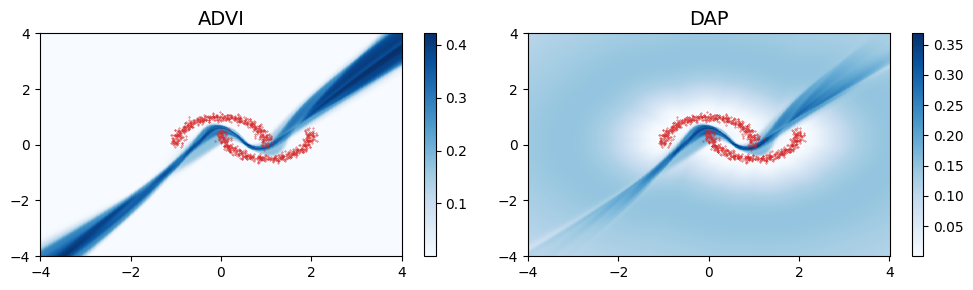}
    \caption{Epistemic uncertainty estimations for two-moon dataset. The left and right columns respectively show estimates produced via ADVI before and after DAP calibration. We can see that DAP calibration helps counteracting ADVI's overconfidence OOD.}
    \label{fig:two_moons}
\end{figure}

We start with two low-dimensional toy examples. The first example is a standard two-moons two-dimensional data set for classification, extensively studied in several works \cite{liu2020simple, van2020uncertainty}. Details about the experimental setup are given in Appendix \ref{app:exp_setups}. We focus on the epistemic variance, as it best characterizes uncertainty OOD. The left panel in Figure \ref{fig:two_moons} shows the epistemic component of the predictive variance produced via Automatic Differentiation Variational Inference (ADVI) \cite{rezende2015variational}. The right panel shows the same after DAP calibration. As we move away from the data, the epistemic uncertainty increase, counteracting the issue of overconfidence OOD.

The second example that we consider is a one-dimensional sinusoidal regression problem with inputs generated either from a Gaussian centered at -5, or from one centered at 5. See Appendix \ref{app:exp_setups} for setup details. Figure \ref{fig:toy_reg} compares predictions and $95\%$ credibility interval estimated via a Laplace approximation \cite{daxberger2021laplace}, before (left) and after (right) DAP calibration. Since we focus on OOD uncertainty, the credibility intervals are produced using epistemic variance only. We can see that DAP calibration reasonably improves uncertainty OOD, remedying the overconfidence issue visible before calibration.
\begin{figure}[t]
    \centering
    \includegraphics[width=\textwidth]{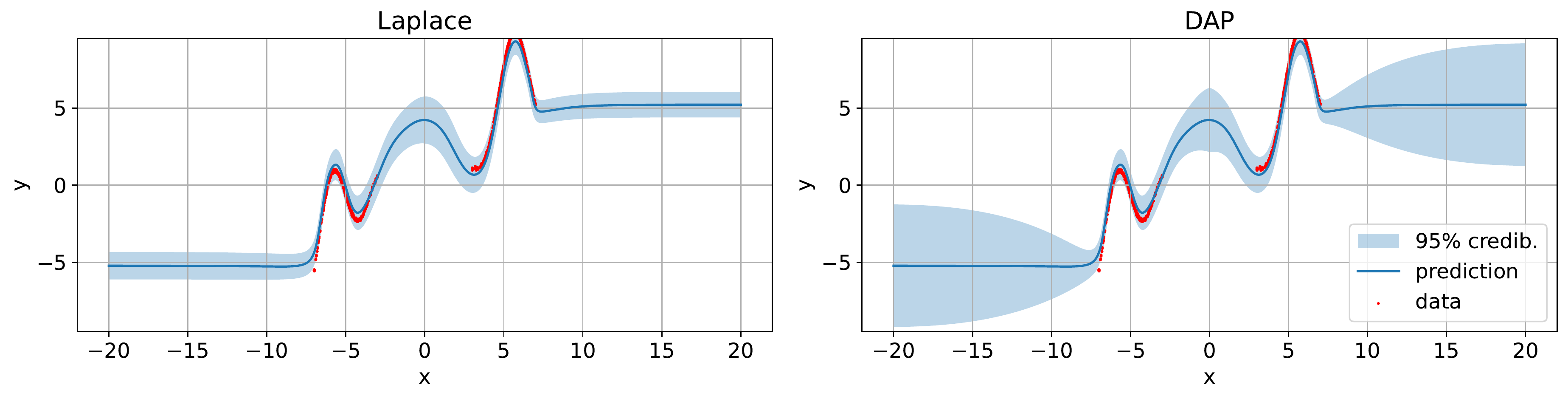}
    \caption{Prediction and 95\% credible interval estimation (via epistemic variance only) using a Laplace approximation, before (left) and after (right) DAP calibration. After DAP calibration, the uncertainty estimates grow as me move away from the data.}
    \label{fig:toy_reg}
\end{figure}

\subsection{UCI-Gap}
\begin{figure}[t]
    \centering
    \includegraphics[width=\textwidth]{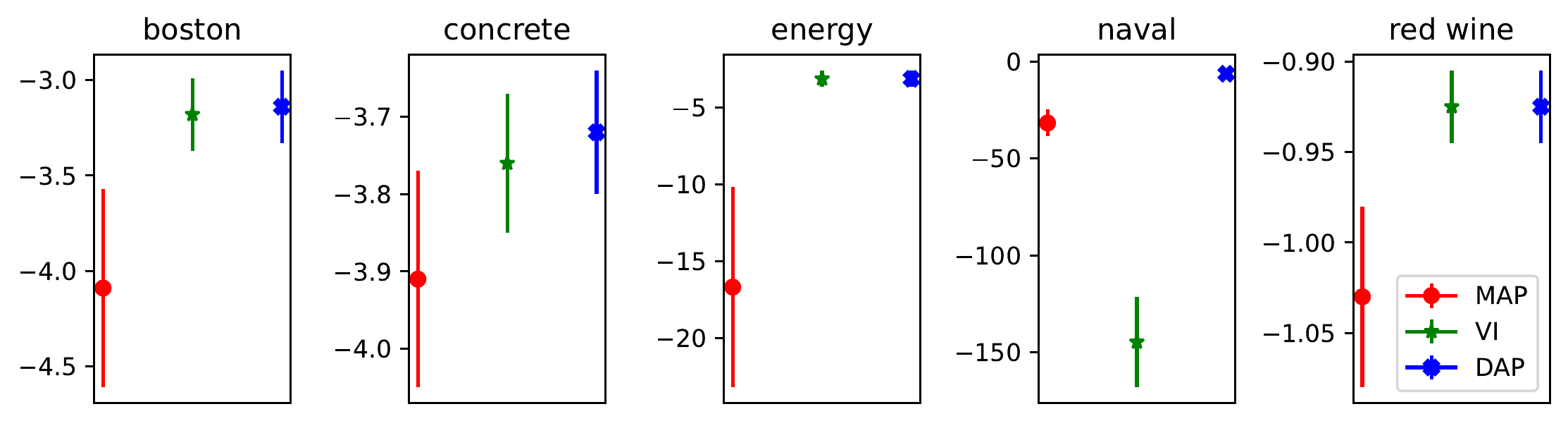}
    \caption{Average test log-likelihood for five UCI "gap" datasets. Results are averaged across different gap splits and displayed within one standard error. The benefit of using DAP calibration is higher when VI is strongly overconfident, like in naval.}
    \label{fig:results_uci}
\end{figure}

We evaluated our method on five UCI "gap" datasets \cite{foong2019between}. The UCI "gap" benchmark has been specifically designed to test for in-between uncertainty, as the test set is obtained by sampling the "middle" region for each input feature component. In Figure \ref{fig:results_uci} we report the average log-likelihood of our method compared MAP estimation and VI. The average is taken among different train-test split. Notice that in this setting the number of splits equals the number of input features.
As it is common in this regression task, we employ a two-layer MLP as our base architecture, and consider its output as the mean of a Gaussian distribution with homoscedastic variance. More details on the experimental setup are provided in the Appendix \ref{app:exp_setups}.
On Boston, concrete, energy and red wine, DAP did not improve much upon black-box VI, indeed it often falls back to the very same solution. On the other hand, as already noticed in \cite{foong2019between}, VI and MAP fail completely on naval due to the large differences between training and test set. Here DAP significantly improves upon the VI solution, correcting its overconfident behaviour.

\subsection{MNIST, Fashion-MNIST and CIFAR-10}
In this section we study the effect of DAP calibration over popular classification datasets, such as MNIST \cite{lecun1998gradient}, Fashion-MNIST \cite{xiao2017fashion} and CIFAR-10 \cite{krizhevsky2009learning}. We compare its performance on wildely used OOD classification benchmarks and compare against: MAP estimation, variational inference (VI) \cite{rezende2015variational}, ODIN \cite{liang2017enhancing}, Deep Deterministic Uncertainty (DDU) \cite{van2020uncertainty} and Deep Ensembles (DE) \cite{lakshminarayanan2017simple}.
% Furthermore, we consider several data sets as OOD, and evaluate the performance of our method on those compared to several benchmarks.

We model target variables via a categorical distribution, with vector of probabilities given by a softmax function evaluated at the outputs of a deep learning model. We employ a simple MLP architecture for MNIST, a LeNet5\cite{lecun1998gradient} for Fashion-MNIST and a a Wide Residual Network 28-10 \cite{zagoruyko2016wide} for CIFAR-10; more details on the experimental setups are provided in the Appendix \ref{app:exp_setups}. After training the models to estimate the MAP, for each model we define the projection $\cP$ of an input as the forward pass evaluated at that input up to the last hidden layer. In order to find the optimal calibration parameter for DAP, we optimize the calibration loss in \eqref{eq:calib_loss} on the misclassified samples on a validation set, as already discussed in Section \ref{sec:calib}. In Appendix \ref{app:distance_mnist}, Figure \ref{fig:MNIST_dist_MMD} confirms the intuition that misclassified inputs tend to be far from the training set, hence it is reasonable to use them for DAP calibration. We highlight that in the following experiments our method is applied on top of the VI solution, but other approximate Bayesian inference methods can alternatively be used.

We measure the area under the receiving operator characteristic curve (AUROC) and the area under the precision recall curve (AUCPR) \cite{davis2006relationship} to quantify how well we discriminate between in-distribution data OOD data sets. This procedure has been employed in several previous works to evaluate the quality of uncertainty estimates OOD \cite{maddox2019simple, liu2020simple, van2020uncertainty, izmailov2021bayesian, mukhoti2021deterministic}. To compute AUROC and AUCPR, we used the predictive mean (confidence score) for DAP, while we rely on predictive entropy for MAP, VI and DE. For DDU we used the Gaussian Mixture Model density as described in \cite{van2020uncertainty}.

\textbf{MNIST and Fashion-MNIST.} In Table \ref{tab:results_mnist} we summarize the results when MNIST is used as the in-distribution dataset and Fashion-MNIST, SVHN and Omniglot are used as out-of-distribution datasets. We observe that DAP performs particularly well on Omniglot and Fashion-MNIST while no major improvement upon VI is noticed for MNIST vs SVHN. We argue that the benefit of DAP calibration is strongest when the posterior inference method is overconfident. Indeed, this is the case for MNIST vs Fashion-MNIST, MNIST vs Omniglot and also for Fashion-MNIST vs MNIST (Table \ref{tab:results_fmnist}) where the performance of VI is close to the one of MAP, which is the weakest among the baselines. We observe that DAP calibration performs either better or competitively with the other methods. In particular, DAP and DDU, which are both distance-aware methods, perform consistently among the best, and better than DE, which is usually considered to be a strong baseline.
% first train the model on the train dataset of MNIST and then compute the uncertainty scores for MNIST test dataset (in-distribution) and different OOD (test) dataset: Fashion-MNIST, Omniglot and SVHN. . The comparison is outlined in Table \ref{tab:results_mnist}, where DAP shows to outperforms other methods on all OOD tasks. Similar conclusions can be drawn from Table \ref{tab:results_fmnist}, where the in-distribution dataset is Fashion MNIST while the out of distribution dataset is MNIST.

\begin{table}[t]
\centering
\caption{Results on MNIST}% using a 2-layer MLP. For DAP $\phi=-8.79$.}
\label{tab:results_mnist}
\resizebox{\columnwidth}{!}{%
\begin{tabular}{cccccccc}
\toprule
\textbf{Model} & \textbf{Accuracy} ($\uparrow$) & \multicolumn{2}{c}{\textbf{F-MNIST}} & \multicolumn{2}{c}{\textbf{SVHN}} & \multicolumn{2}{c}{\textbf{Omniglot}} \\ \cline{3-8} 
                                &                                                                           & \textbf{AUROC} ($\uparrow$)  & \textbf{AUCPR} ($\uparrow$) & \textbf{AUROC} ($\uparrow$)   & \textbf{AUCPR} ($\uparrow$) & \textbf{AUROC} ($\uparrow$) & \textbf{AUCPR} ($\uparrow$) \\ \midrule
MAP                                         & 96.40                             & 87.65           & 87.77           & 92.38             & 95.26  & 89.58 & 90.65             \\
\midrule
ODIN                                        &  -                           & 89.50                & 86.42                & 97.10                  &  94.85     & 90.15 & 87.49
\\
\midrule
DDU                                  & -                             & 94.35           & 95.53           & 91.41             & 90.07  & 92.85 & 92.66          \\
\midrule
DE                                  & 97.24                             & 90.56           & 88.82           & 93.09             & 96.54  & 94.70 & \textbf{95.23}          \\
\midrule
VI                                              & 96.62                            & 87.90                & 89.76                & 98.46                  &  \textbf{98.96}   & 89.96 & 91.06  
\\

\midrule
DAP                                                 & 96.57                                  &  \textbf{96.25}               & \textbf{96.44}                & \textbf{98.85}                  & 97.99    & \textbf{95.24} & 95.14 \\ \bottomrule              
\end{tabular}
}
\end{table}

\begin{table}[t]
\centering
\makebox[0pt][c]{\parbox{1.\textwidth}{%
\begin{minipage}[b]{0.39\hsize}\centering
    % \begin{table}
    % \centering
    \caption{Results on Fashion-MNIST} %using a CNN. For DAP $\phi=-10.45$} 
    \label{tab:results_fmnist}
    \resizebox{\columnwidth}{!}{%
    % \begin{tabular}{cccc}
    % \toprule
    % \multirow{\textbf{Model}}  & \multirow{\textbf{Accuracy ($\uparrow$)}} & \multicolumn{2}{c}{\textbf{MNIST}} \\ \cline{3-4} 
    %                                                               &                            & \textbf{AUROC ($\uparrow$)}  & \textbf{AUCPR ($\uparrow$)} \\ \midrule
    % MAP                                                 & 91.28                              & 80.44           & 78.60 \\
    % \midrule
    % ODIN                                              &                          & 82.53           & 82.51 \\
    % \midrule
    % DDU                                                    & 91.28                             & 92.29           & 94.50 \\
    % \midrule
    % DE                                                    & 91.63                             & 89.63           & 89.58 \\
    % \midrule
    % VI                                       & 91.72                              & 87.24           & 88.09 \\
    % \midrule
    % DAP                              & 91.72                                 & \textbf{98.31}           & \textbf{98.69} \\ 
    % \bottomrule              
    % \end{tabular}
        \begin{tabular}{cccc}
    \toprule
\textbf{Model} & \textbf{Accuracy ($\uparrow$)} & \multicolumn{2}{c}{\textbf{MNIST}}                      \\ \cline{3-4} 
                                &                                                 & \textbf{AUROC ($\uparrow$)} & \textbf{AUCPR ($\uparrow$)} \\
\midrule
MAP                             & 89.39                                           & 67.65                      & 67.49                      \\
\midrule
ODIN                            &   -                                              & 71.22                      & 64.79                      \\
\midrule
DDU                             &   -                                              & 99.66                      & 99.73                      \\
\midrule
DE                              & 91.34                                           & 91.69                      & 91.92                      \\
\midrule
VI                              & 89.66                                           & 70.31                      & 71.98                      \\
\midrule
DAP                             & 89.66                                           & \textbf{99.70}                      & \textbf{99.76}                      \\
\bottomrule
\end{tabular}
    }

    % \end{table}
\end{minipage}
\hspace{0.1cm}
\begin{minipage}[b]{0.59\hsize}\centering
%     \begin{table}
% \centering
\caption{Results on CIFAR-10} % using a WideResNet 28-10. For DAP $\phi=-14.37$.}
\label{tab:results_CIFAR-10}
\resizebox{\columnwidth}{!}{%
\begin{tabular}{ccccccc}
\toprule
\textbf{Model} & \textbf{Accuracy} ($\uparrow$) & \multicolumn{2}{c}{\textbf{SVHN}} & \multicolumn{2}{c}{\textbf{CIFAR-100}} \\ \cline{3-6} 
                                &                               & \textbf{AUROC ($\uparrow$)}  & \textbf{AUCPR ($\uparrow$)}  & \textbf{AUROC ($\uparrow$)}    & \textbf{AUCPR ($\uparrow$)}    \\ \midrule
MAP                              & 95.40                           & 94.91           & 97.16           & 88.41             & 86.70             \\
\midrule
ODIN                               &  -                            & 94.55           & 92.12           & 88.26             & 88.20             \\
\midrule
DDU                             & -                         & \textbf{97.04}           & 94.91           & 90.60             & 91.13             \\
\midrule
DE                             & 96.14                         & 95.79           & 97.72           & 90.94             & 91.01             \\
\midrule
VI                                   & 95.24                           & 96.73               & \textbf{98.33}                & 90.70                  &  88.68   \\
\midrule
DAP                                     & 95.24                             & 95.24               & 94.23                & \textbf{90.95}                & \textbf{91.87}                      \\ \bottomrule               
\end{tabular}
}
% \end{table}
\end{minipage}%
}}
\end{table}
\textbf{Cifar 10.} We train the model on CIFAR-10 and then consider a standard OOD task using SVHN (far-OOD detection) and a more complex OOD task using CIFAR-100 (near-OOD detection). Table \ref{tab:results_CIFAR-10} shows that our model is competitive with all the baselines, including DDU and DE, on both tasks. Our model is the top performer on the harder task of far-OOD detection (CIFAR-100) while it falls shortly behind other methods on CIFAR-10 vs SVHN. As observed earlier, in the latter task the VI model, whose posterior samples DAP builds upon in this experiment, is already achieving remarkable results, and it is possible that it does not exist a calibration parameter $\phi$ that allows to improve upon the VI solution. We remark that, in order to improve DAP calibration scores, we could have calibrated individually for each of the OOD data sets. However, we preferred to stay agnostic to the OOD data set and calibrate $\phi$ only once over the misclassified validation data, as described above.

% Please add the following required packages to your document preamble:
% \usepackage{multirow}
% \begin{table}[h!]
% \centering
% \caption{Average test log-likelihood for five UCI "gap" datasets. Results are averaged across different gap splits (the number of splits is the number of input features for each dataset) and displayed within one standard error.}
% \label{tab:results_uci}
% \begin{tabular}{cccccc}
% \toprule
% \multirow{2}{*}{\textbf{Model}} & \multicolumn{5}{c}{\textbf{Dataset}}                \\ \cline{2-6} 
%                                 & \textbf{boston}  & \textbf{concrete} & \textbf{energy} & \textbf{naval}    & \textbf{wine}                    \\ 
% \midrule
% MAP                             & $-4.09\pm0.52$   & $-3.91\pm0.14$    & $-16.69\pm6.53$ & $-31.71\pm6.89$   & $-1.03\pm0.05$                   \\
% ADVI                            & $-3.18 \pm 0.19$ & $-3.76\pm0.09$    & $-3.12\pm0.54$  & $-144.71\pm23.36$ & $-0.925 \pm 0.02$                \\
% DAP                             & $-3.14 \pm 0.19$ & $-3.72\pm0.08$    & $-3.12\pm0.54$  & $-6.31\pm1.69$    & $-0.925 \pm 0.02$ \\
% \bottomrule
% \end{tabular}
% \end{table}

\section{Conclusion and future directions}
In this work we introduced a type of OOD uncertainty calibration based on distance-aware priors (DAPs). These prior distributions assume that model parameters depend on the inputs through a measure of their distance from the training data. DAP calibration can be done as a post-processing step of posterior inference, and it can be easily integrated with other popular calibration techniques. We performed extensive evaluation on both classification and regression tasks, demonstrating competitive performance against several other methods that aim to estimate OOD uncertainty. We believe DAP can be used as a practical option to decrease the issue of overconfidence in out-of-distribution settings. We expect this approach to have broad impact, since in virtually any real-world problem our test points are not drawn from exactly the same distribution as our training points, and we must have carefully calibrated uncertainty for decision making in these settings.

A natural extension of this work is to couple DAP calibration with other methods for in-domain calibration, such as Platt scaling \cite{platt1999probabilistic}. Furthermore, it would be interesting to investigate the impact of feature-collapse \cite{van2020uncertainty,liu2020simple,mukhoti2021deterministic,van2021on} when using neural networks as the projector in the distance computation in 
DAP. Another important direction is to improve the importance sampling approach by employing variance reduction strategies that can mitigate its known instability issues \cite{sutton2018reinforcement}.

%In this work we did not consider the issue of feature collapse when using neural networks as feature extractor. This problem has been recently studied in the literature  and multiple solutions that help mitigate this issue have already been proposed. Thus, a natural extension of this work is to investigate the impact of feature-collpase in DAP and quantify the benefit of coupling it with methods such as Spectral Normalization. An other interesting future research direction is to combine DAP, which focuses on epistemic uncertainty calibration, with other methodologies tailored to calibrate aleatoric uncertainty. Furthermore, importance sampling is a compelling choice

\newpage
\printbibliography
\newpage

\section*{Checklist}
\begin{enumerate}

\item For all authors...
\begin{enumerate}
  \item Do the main claims made in the abstract and introduction accurately reflect the paper's contributions and scope?
    \answerYes{}
  \item Did you describe the limitations of your work?
    \answerYes{We discussed that the scope of our work is limited to OOD uncertainty calibration.}
  \item Did you discuss any potential negative societal impacts of your work?
    \answerNo{We do not see any.}
  \item Have you read the ethics review guidelines and ensured that your paper conforms to them?
    \answerYes{}
\end{enumerate}

\item If you are including theoretical results...
\begin{enumerate}
  \item Did you state the full set of assumptions of all theoretical results?
    \answerYes{}
        \item Did you include complete proofs of all theoretical results?
    \answerYes{}
\end{enumerate}

\item If you ran experiments...
\begin{enumerate}
  \item Did you include the code, data, and instructions needed to reproduce the main experimental results (either in the supplemental material or as a URL)?
    \answerNo{We do not yet provide the code, but we will after the papers is accepted.}
  \item Did you specify all the training details (e.g., data splits, hyperparameters, how they were chosen)?
    \answerYes{}
        \item Did you report error bars (e.g., with respect to the random seed after running experiments multiple times)?
    \answerNo{}
        \item Did you include the total amount of compute and the type of resources used (e.g., type of GPUs, internal cluster, or cloud provider)?
    \answerNo{Not relevant for this paper, as we do not compare computational times.}
\end{enumerate}

\item If you are using existing assets (e.g., code, data, models) or curating/releasing new assets...
\begin{enumerate}
  \item If your work uses existing assets, did you cite the creators?
    \answerNA{}
  \item Did you mention the license of the assets?
    \answerNA{}
  \item Did you include any new assets either in the supplemental material or as a URL?
    \answerNA{}
  \item Did you discuss whether and how consent was obtained from people whose data you're using/curating?
    \answerNA{}
  \item Did you discuss whether the data you are using/curating contains personally identifiable information or offensive content?
    \answerNA{}
\end{enumerate}

\item If you used crowdsourcing or conducted research with human subjects...
\begin{enumerate}
  \item Did you include the full text of instructions given to participants and screenshots, if applicable?
    \answerNA{}
  \item Did you describe any potential participant risks, with links to Institutional Review Board (IRB) approvals, if applicable?
    \answerNA{}
  \item Did you include the estimated hourly wage paid to participants and the total amount spent on participant compensation?
    \answerNA{}
\end{enumerate}

\end{enumerate}

%%%%%%%%%%%%%%%%%%%%%%%%%%%%%%%%%%%%%%%%%%%%%%%%%%%%%%%%%%%%

\newpage
\appendix

\section{The posterior weight}\label{app:post_weight}
The relation between the distance-aware prior distribution $p(\theta|x_j^*,\bx)$ and the posterior distribution $p(\theta|x_j^*,\bx, \by)$ in \eqref{eq:pred_distr} is given by Bayes' theorem:
\[
    p(\theta|x_j^*,\bx,\by) = \frac{p(\by|\theta,\bx)p(\theta|x_j^*,\bx)}{p(\by|x_j^*,\bx)},
\]
where $p(\by|\theta,\bx)$ is the training likelihood, $p(\theta|x_j^*,\bx)$ is the distance-aware prior and $p(\by|x_j^*,\bx)$ is the marginal training likelihood, i.e. the evidence. Then, by using Bayes' theorem on both $p(\theta|\bx,\by)$ and $p(\theta|x_j^*,\bx,\by)$, we get
\begin{align*}
     w(\theta|x_j^*,\bx,\by)&:=\frac{p(\theta|x_j^*,\bx,\by)}{p(\theta|\bx,\by)} = \frac{p(\theta|x_j^*,\bx)}{p(\theta|\bx)}\frac{p(\by|\bx)}{p(\by|x_j^*,\bx)}. 
\end{align*}

\section{The marginal ratio in the Gaussian scenario}\label{app:gauss_marginal}
We first note that the expected value of $w(\theta|x_j^*,\bx,\by)$ with respect to $p(\theta|x_j^*,\bx)$ is one:
\[
\bE_{\theta|\bx,\by}[w(\theta|x_j^*,\bx,\by)] = \int \frac{p(\theta|x_j^*,\bx,\by)}{p(\theta|\bx,\by)}p(\theta|\bx,\by)\,d\theta = \int p(\theta|x_j^*,\bx,\by)\,d\theta = 1. \]
Given the expression of posterior weight (see Appendix \ref{app:post_weight}), this implies
\[ \frac{p(\by|\bx)}{p(\by|x_j^*,\bx)} = \bE_{\theta|\bx,\by}\left[\frac{p(\theta|x_j^*,\bx)}{p(\theta|\bx)}\right]^{-1}. \]

Secondly, let us assume
\begin{align*}
    p(\theta|x_j^*,\bx) &:= N(\theta|0, (\sigma_0+d_j)^2I_{n_\theta}),\\
    p(\theta|\bx) &:= N(\theta|0, \sigma_0^2I_{n_\theta}),\\
    p(\theta|\bx,\by) &:= N(\theta|\mu, \diag(\sigma^2)),
\end{align*}
where $d_j:=d([x_j^*,\bx])$, and $\diag(\sigma^2)$ is a diagonal matrix with $\sigma^2\in\bR_+^{n_\theta}$ on the diagonal. We have
\[ \frac{p(\theta|x_j^*,\bx)}{p(\theta|\bx)} = \left(\frac{\sigma_0}{\sigma_0+d_j}\right)^{n_\theta} \exp\left(-\frac{\|\theta\|^2}{2}\left(\frac{1}{(\sigma_0+d_j)^2} - \frac{1}{\sigma_0^2}\right)\right) = \left(\frac{\sigma_0}{\sigma_0+d_j}\right)^{n_\theta}\exp\left(\frac{\|\theta\|^2}{2\gamma_j^2}\right), \] 
with
\[ \frac{1}{\gamma_j^2} := \frac{1}{\sigma_0^2} - \frac{1}{(\sigma_0+d_j)^2}.  \] 
Let us define $\beta_{i,j}^{-2} :=\frac{1}{\sigma_i^2} - \frac{1}{\gamma_j^2}$, and assume $\beta_{i,j}^2 > 0$. We will show later when this condition is true. Then
\begin{align*} 
&\bE_{\theta|\bx,\by}\left[\frac{p(\theta|x_j^*,\bx)}{p(\theta|\bx)}\right] =\frac{1}{(2\pi)^{n_\theta/2}} \left(\frac{\sigma_0}{\sigma_0+d_j}\right)^{n_\theta} \prod_i\frac{1}{\sigma_i}\int \exp\left(-\frac{1}{2}\left(\frac{\theta_i^2 - 2\theta_i\mu_i}{\sigma_i^2} -\frac{\theta_i^2}{\gamma_j^2} + \frac{\mu_i^2}{\sigma_i^2}\right)\right)\,d\theta_i\\
&=\frac{1}{(2\pi)^{n_\theta/2}} \left(\frac{\sigma_0}{\sigma_0+d_j}\right)^{n_\theta} \prod_i\frac{1}{\sigma_i}\int \exp\left(-\frac{1}{2}\left(\left(\frac{1}{\sigma_i^2} - \frac{1}{\gamma_j^2}\right)\theta_i^2 - 2\theta_i\frac{\mu_i}{\sigma_i^2} + \frac{\mu_i^2}{\sigma_i^2}\right)\right)\,d\theta_i\\
&=\frac{1}{(2\pi)^{n_\theta/2}} \left(\frac{\sigma_0}{\sigma_0+d_j}\right)^{n_\theta} \prod_i\frac{1}{\sigma_i}\int \exp\left(-\frac{1}{2}\left(\beta_{i,j}^{-2}\left(\theta_i^2 - 2\theta_i\beta_{i,j}^2\frac{\mu_i}{\sigma_i^2} + \beta_{i,j}^4\frac{\mu_i^2}{\sigma_i^4}\right) - \beta_{i,j}^2\frac{\mu_i^2}{\sigma_i^4} + \frac{\mu_i^2}{\sigma_i^2}\right)\right)\,d\theta_i\\
&=\frac{1}{(2\pi)^{n_\theta/2}} \left(\frac{\sigma_0}{\sigma_0+d_j}\right)^{n_\theta} \prod_i\frac{1}{\sigma_i}\exp\left(-\frac{1}{2}\frac{\mu_i^2}{\sigma_i^2}\left(1 - \frac{\beta_{i,j}^2}{\sigma_i^2}\right)\right)\int \exp\left(-\frac{1}{2\beta_{i,j}^2}\left(\theta_i - \beta_{i,j}^2\frac{\mu_i}{\sigma_i^2}\right)^2\right)\,d\theta_i\\
&= \left(\frac{\sigma_0}{\sigma_0+d_j}\right)^{n_\theta} \prod_i\frac{\beta_{i,j}}{\sigma_i}\exp\left(-\frac{1}{2}\frac{\mu_i^2}{\sigma_i^2}\left(1 - \frac{\beta_{i,j}^2}{\sigma_i^2}\right)\right).
\end{align*}
Because of the identity shown above, we have 
\[ \frac{p(\by|\bx)}{p(\by|x_j^*,\bx)} = \left(\frac{\sigma_0+d_j}{\sigma_0}\right)^{n_\theta} \prod_i\frac{\sigma_i}{\beta_{i,j}}\exp\left(\frac{1}{2}\frac{\mu_i^2}{\sigma_i^2}\left(1 - \frac{\beta_{i,j}^2}{\sigma_i^2}\right)\right). \]

Let us now define $\bsigma^2:=\max(\sigma_i^2)$. We show that if $\bsigma^2<\sigma_0^2$, then $\beta_{i,j}^2>0$ for all $i=1,\dots,n_\theta$. Let us start by the thesis:
\[ \frac{1}{\sigma_i^2} - \frac{1}{\gamma_j^2} > 0\quad\quad\forall i=1,\dots,n_\theta, \]
which is verified if and only if
\[ \gamma_j^2 > \max(\sigma_i^2)=\bsigma^2. \] 
By definition of $\gamma_j^2$, this is true if and only if
\[ \sigma_0^2(\sigma_0+d_j)^2 > \bsigma^2(d_j^2 +2d_j\sigma_0). \]
By developing the products, we get
\[ \sigma_0^4 + \sigma_0^2d_j^2 + 2\sigma_0^3d_j > \bsigma^2d_j^2 + 2\sigma_0\bsigma^2d_j. \]
Finally, by factorization, we have
\[ (\sigma_0^2-\bsigma^2)d_j^2 + 2\sigma_0(\sigma_0^2-\bsigma^2)d_j + \sigma_0^4>0, \]
which is clearly always positive given the assumption $\bsigma^2<\sigma_0^2$.

\section{Experimental setups}\label{app:exp_setups}
\paragraph{Toy classification.} We generate 1000 samples with noise given by 0.07. We model inputs to logits via a simple multilayer perceptron (MLP) and take a binomial distribution over the targets, with probabilities given by the sigmoid function of the logits. As a prior distribution over the model parameters, we take a standard Gaussian. As validation inputs $x_j^*$, we used samples from Gaussian distribution centered just outside the training domain, with standard deviation given by $0.1$. We take the pre-distance in \eqref{eq:d0_nn} and the calibration loss in \eqref{eq:calib_loss}, with $\cU_\phi(x_j^*)$ being the predictive mean and $\gamma=[0.5, 0.5]$. Results are plotted for $\phi=-4.85$, that is a minimum of the calibration loss.

\paragraph{Toy regression.}
 We model the target distribution via a Gaussian, where mean and standard deviations are modelled via independent MLPs. We use a standard Gaussian over all model parameters. Since only the parameters of the mean model directly contribute to the epistemic uncertainty, we do DAP calibration only on them, that is we define a DAP that increases the prior variance with a distance only for the components of $\theta$ corresponding for the parameters of the mean model. This helps better isolating the effect that we want to achieve,  that is increasing the epistemic uncertainty OOD without necessarily modifying the aleatoric uncertainty. We employ the pre-distance in \eqref{eq:d0_nn} and the calibration loss in \eqref{eq:calib_loss}, where we take $\cU_\phi(x_j^*)$ to be the epistemic variance, and $\gamma=10 q_{95}$, with $q_{95}$ being the 95th of the predicted epistemic variances at the training inputs. The validation inputs are simulated from three Gaussians centered just outside of the training domain, on the left, in between and on the right of the clusters. 

\paragraph{UCI Gap.} 
We used a minibatch of size equal to 64, and optimize hyperparameters via grid search on a validation set held out from the training set (90-10 split). We highlight that this validation set is later used for DAP calibration and not included in the final training of the model. 
For MAP and VI the hyperparameters that we optimised were the learning rates [1e-2,1e-4, 1e-5] and the number of epochs [20,40,100,200], while we kept the prior variance fixed to 1. We used the Adam \cite{kingma2014adam} optimizer. For VI we used an ensemble of 100 posterior samples at prediction time. The calibration parameter $\phi$ of DAP was tuned independently for each split using a subsample of the validation set used for hyperparameter tuning. This subset includes the samples used for calibration (30\% of the total), where the model has the higher RMSE. Like in the toy regression example, we employ the pre-distance in \eqref{eq:d0_nn} and the calibration loss in \eqref{eq:calib_loss}, where we take $\cU_\phi(x_j^*)$ to be the epistemic variance, and $\gamma=10 q_{95}$, with $q_{95}$ being the 95th of the predicted epistemic variances at the training inputs.

\paragraph{Classification experiments.} We employed a two-layer MLP with 30 hidden layers and tanh activations for the MNIST experiments. We trained the model for 50 epochs with SGD with momentum of 0.9, a learning rate of $10^{-6}$ and batch size of 128. For VI we used 50 posterior samples at inference time. The data are normalized between 0 and 1. For SVHN and Omniglot, the images are turned to gray scale and resized to have height and width of 28 before being normalized.

For Fashion-MNIST, we trained a LeNet5 for 80 epochs with SGD with momentum of 0.9, a learning rate of $10^{-6}$ and batch size of 512. For VI we used 50 posterior samples at inference time. The data are normalized between 0 and 1. 

We employed a WideResNet with depth 28 and widen factor 10 for the CIFAR-10 experiments. The network was trained for 200 epochs without early stopping and batch size of 128 decaying the learning rate by a factor of 0.2 every 60 epochs. The optimizer is SGD with momentum of 0.9 and an initial learning rate of 0.1. We initialized the mean of VI with a pre-trained MAP network and then used 20 posterior samples at evaluation time.

For ODIN and DDU we used MAP as the pretrained feature extractor. In DAP, we also used the MAP model to transform the inputs before feeding them to the distance function in order to ease the comparison with the other methods. For ODIN we chose the temperature parameter $T$ among $[1,10,100,300,500,1000]$ and the noise parameter $\epsilon$ within $(0.001, 0.5)$ with step size 0.017. For deep ensembles, we used always 5 networks. We use an hold-out of 1000 samples from the in-distribution data for calibrating DAP. As described in Section \ref{sec:experiments}, among these samples only the misclassified ones are used in practice in the calibration procedure.

\section{Distance plot on MNIST}\label{app:distance_mnist}
\begin{figure}[h!]
    \centering
    \includegraphics[width=0.5\textwidth]{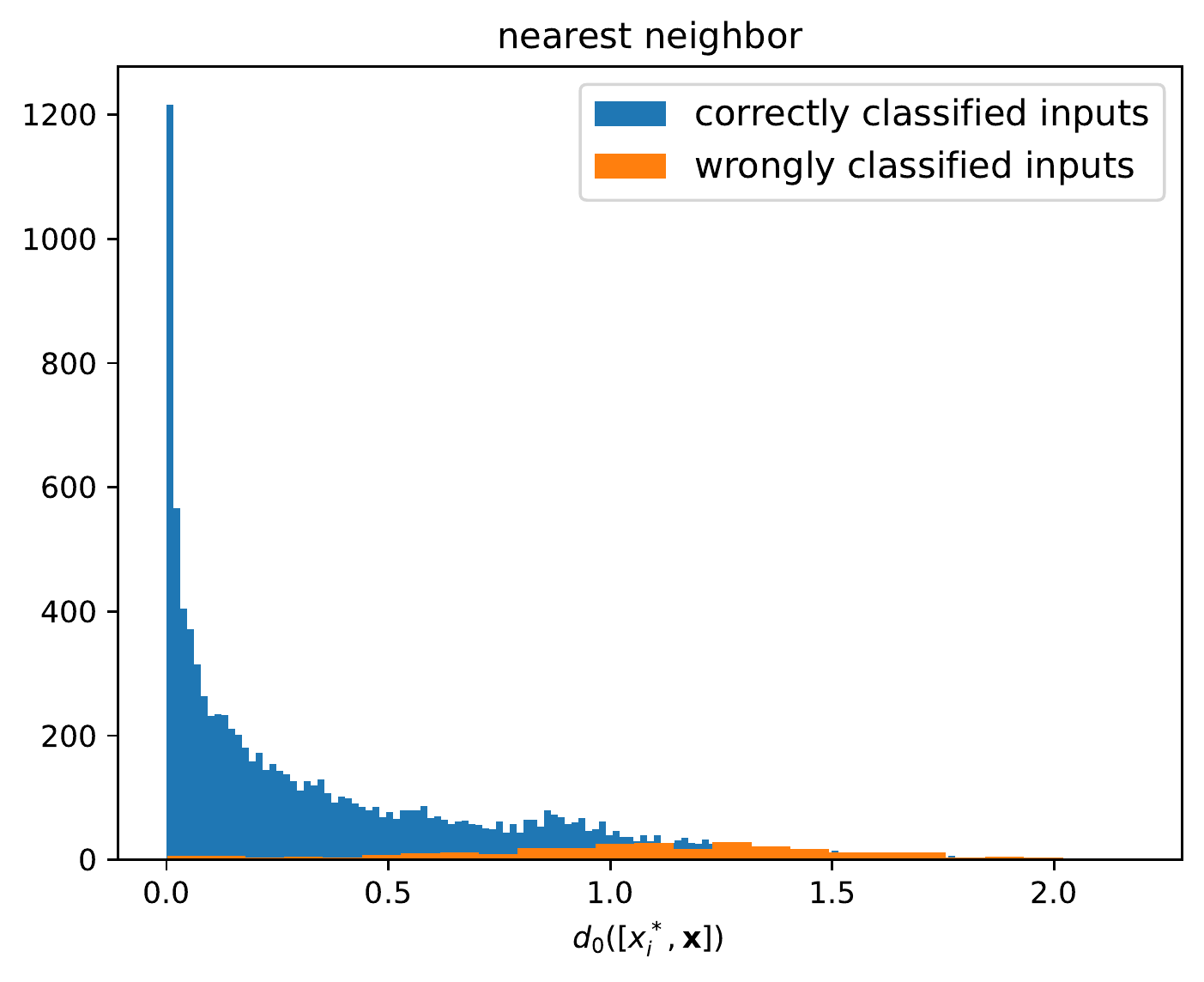}
    \caption{Distances from the training set, for inputs correctly and wrongly classified. The plots confirm the intuition that wrongly classified inputs tend to be further from the training domain.}
    \label{fig:MNIST_dist_MMD}
\end{figure}
\newpage
\section{Calibration losses for MNIST, Fashion-MNIST and CIFAR-10}\label{app:calib_losses}
\begin{figure}[h!]
    \centering
    \includegraphics[width=\textwidth]{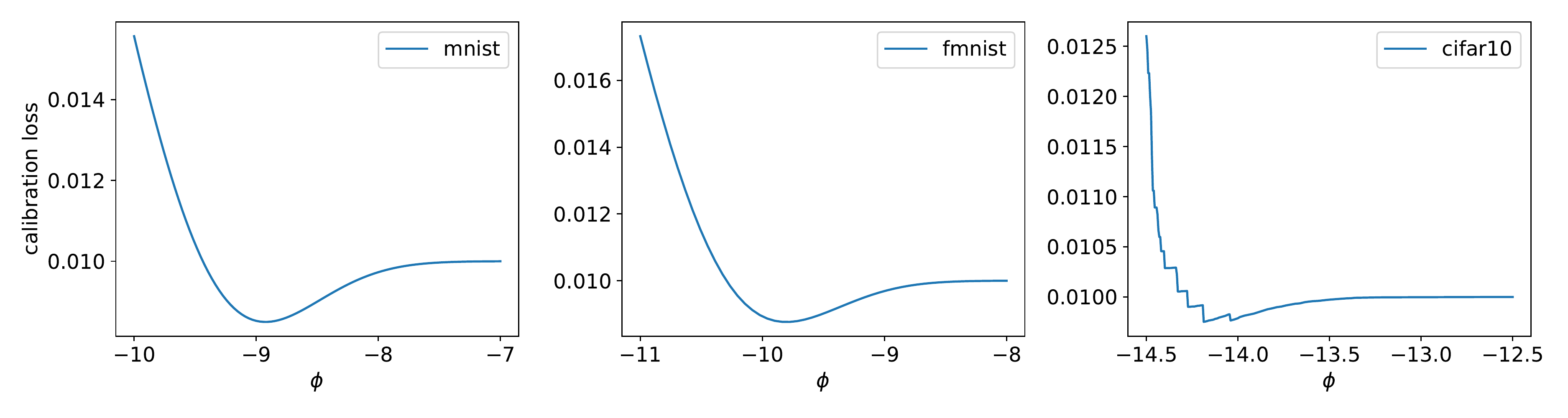}
    \caption{Calibration losses over $\phi$ for MNIST, Fashion-MNIST and CIFAR-10. The plots visualize where the losses achieve their minima.}
    \label{fig:calib_losses}
\end{figure}

\end{document}